\documentclass{article}
\usepackage{spconf,amsmath,amsfonts,graphicx,verbatim,booktabs,color, algorithm, algorithmic, subfigure, enumitem}
\usepackage{amsmath,bm}
\usepackage{stfloats}
\usepackage{setspace}
\usepackage[nobreak]{cite}
\usepackage{multirow}
\usepackage{array}
\usepackage{url}

\title{Prompt-based Personalized Federated Learning\\ for Medical Visual Question Answering}
\name{He Zhu$^{\dag}$, Ren Togo$^{\dag\dag}$, Takahiro Ogawa$^{\dag\dag}$ and Miki Haseyama$^{\dag\dag}$\thanks{This work was partly supported by JSPS KAKENHI Grant Number JP21H03456 and JP23K11141.}}
\address{
    $^{\dag}$Graduate School of Information Science and Technology, Hokkaido University, Japan\\
    $^{\dag\dag}$Faculty of Information Science and Technology, Hokkaido University, Japan\\
    E-mail: \{zhu, togo, ogawa, mhaseyama\}@lmd.ist.hokudai.ac.jp
    }

\begin{document}
\ninept
\maketitle
%

\begin{abstract}
We present a novel prompt-based personalized federated learning (pFL) method to address data heterogeneity and privacy concerns in traditional medical visual question answering (VQA) methods. Specifically, we regard medical datasets from different organs as clients and use pFL to train personalized transformer-based VQA models for each client. To address the high computational complexity of client-to-client communication in previous pFL methods, we propose a succinct information sharing system by introducing prompts that are small learnable parameters. In addition, the proposed method introduces a reliability parameter to prevent the negative effects of low performance and irrelevant clients. Finally, extensive evaluations on various heterogeneous medical datasets attest to the effectiveness of our proposed method.
\end{abstract}
\section{Introduction}
\label{sec:intro}
Rising poverty levels and an increasing aging population are exacerbating the global shortage of healthcare resources. Medical AI is now seen as a pivotal solution to address these problems in the foreseeable future~\cite{yu2018artificial}.
The general tasks in medical AI, such as classification~\cite{7930302} and segmentation~\cite{masood2015survey}, lack adaptability to rapidly changing clinical situations and often neglect patient involvement in medical diagnoses.
The recent rise of chatGPT~\cite{openai_gpt3} demonstrates capabilities that closely resemble human conversation and analytical skills, showing the potential of the language for medical applications.
As a vision-language task, medical visual question answering (VQA)~\cite{lin2023medical} can interpret questions framed in natural language and provide accurate, user-friendly responses. This is expected to alleviate the workload on physicians by facilitating communication with patients.

While previous research has primarily focused on improving model accuracy~\cite{gong2021cross}, two significant challenges remain in the medical domain. First, data distributions of the different patients are difficult to learn by a single model due to the heterogeneity such as feature distribution drift and label distribution skew~\cite{schork2019artificial}. The other is that due to patient privacy concerns, the volume of data available for training remains constrained~\cite{wu2020personalized}. 
In the general domain, the federated learning (FL)~\cite{xu2021federated} and the personalized federated learning (pFL)~\cite{tan2022towards} are proposed for training personalized models separately using private data.
FL utilizes a central server to manage the global model, enabling certain clients to participate.
On the contrary, pFL has been developed to learn a customized model for each client that performs better on private data while benefiting from collaborative training.
Essentially, each pFL client model is uniquely calibrated to its respective user, and the individual clients do not communicate any data during the training process. However, due to the limitations of the previous pFL framework, it is challenging to introduce it directly into medical VQA.

The traditional pFL framework~\cite{zhu2023confidence} allows for information sharing by communicating the model weights of each client~\cite{kulkarni2020survey}, which leads to many parameters requiring updates. As a result of computational limitations, pFL applications have been mostly confined to Deep Neural Networks (DNNs) that handle simpler datasets such as MNIST~\cite{deng2012mnist} and CIFAR10/100~\cite{krizhevsky2009learning}.
However, the medical VQA task has textual information as an additional modality and typically deals with informative images. Furthermore, the transformer model~\cite{vaswani2017attention}, known for its state-of-the-art performance in the VQA task, inherently possesses more parameters than conventional DNNs. Although Li et al.~\cite{li2023fedtp} proposed the personalized simple transformer with few blocks~\cite{vaswani2017attention}, it follows the previous consumptive parameter communication, which brings a massive consumption of computing resources. Inspired by Wang et al.~\cite{wang2022learning, wang2022dualprompt}, we introduce prompt learning~\cite{liu2023pre} into the pFL method to reduce the number of parameters involved in client communication, which can reduce the computational complexity by replacing model weights with prompts.

In this paper, we propose prompt-based personalized federated learning for medical VQA to address the heterogeneity and privacy problem.
We establish clients for heterogeneous medical data from different organs and simultaneously train personalized VQA models for each client locally, where each model can be regarded as learning a different data distribution.
To achieve privacy protection, each client can only access its own private data during training, and information is shared among clients through model parameters.
Specifically, we introduce the prompt, a small learnable parameter with high dimensionality, as an additional input to the residual attention blocks in the transformer encoder. We only update the prompt during training so that it obtains the information in the training data. 
Since a client's prompts are accessible to other clients, we facilitate client-to-client communication by integrating the prompts to generate shared information. 
Moreover, we propose a reliability parameter that determines the weight of each client in the communication process.
Our experimental findings indicate that parameter communication between clients outperforms individualized training.
Finally, we summarize our contributions of this paper as follows.
\begin{itemize}[itemsep=0pt, parsep=0pt,topsep=0pt,partopsep=0pt]
\item We propose a novel prompt-based pFL method for personalized medical VQA addressing the problems of the heterogeneity and privacy preservation in medical domain.
\item We provide a simple method to represent model information using prompts, which also reduces the computational complexity of the client-to-client communication process.
\item We propose a reliability parameter for controlling the weight of each client in the communication process to avoid negative effects.
\end{itemize}
\section{PROMPT-BASED PERSONALIZED FEDERATED LEARNING FOR MEDICAL VQA}
\begin{figure}[t]
\centering   
\includegraphics[width=8.5cm]{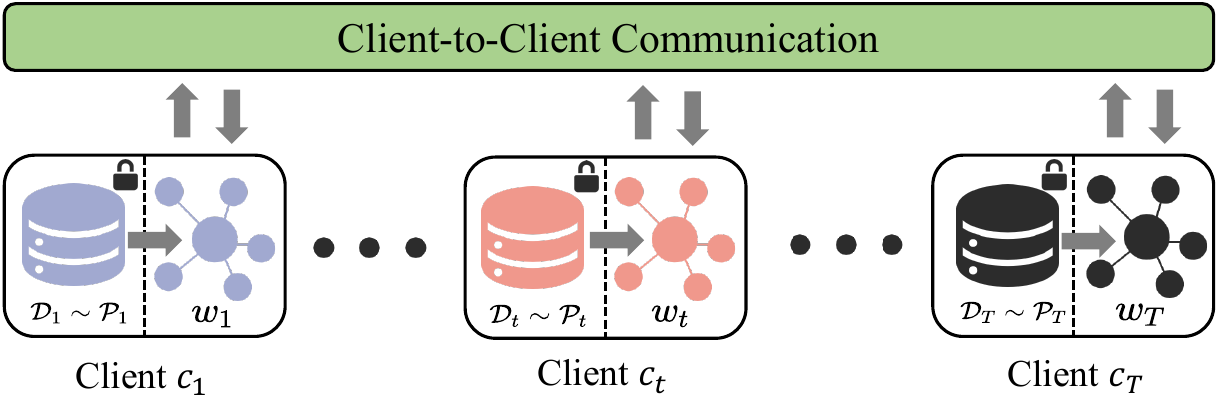}
\caption{An overview of the proposed prompt-based pFL method. A client $c_t$ has private data $\mathcal{D}_t$ following the distribution $\mathcal{P}_t$, and a personalized client VQA model with the weight $\bm{w}_t$. The client uploads the local information and obtains shared information through client-to-client communication.}
\label{fig:overview}
\vspace{-1em}
\end{figure}
We provide an overview of our method in Fig.~\ref{fig:overview}. We first formalize the proposed prompt-based pFL problem and then present the pFL medical VQA method.

\subsection{Problem Formulation}
The proposed method aims to train personalized medical VQA models for $T$ clients collaboratively. Each client has its own private data and can communicate with others without sharing these data.
The $t$-th client denoted as $c_t$ is characterized by its data distribution $\mathcal{P}_t: \mathcal{X} \times \mathcal{A}$, where $\mathcal{X}$ represents the input data space and $\mathcal{A}$ denotes the answer space. Assume that client $c_t$ has access to $N_t$ independent and identically distributed (IID) samples from $\mathcal{P}_t$, denoted by $\mathcal{D}_t = \lbrace(I_t^j, Q_t^j, A_t^j)\rbrace_{j=1}^{N_t}$, where $I_t^j$, $Q_t^j$ and $A_t^j$ respectively represent the input images, questions and the ground truth answers.

Let $\mathcal{L}_t: \mathcal{X} \times \mathcal{A} \rightarrow \mathbb{R}+$ represent the loss function associated with client $c_t$ and the data sampled from the distribution $\mathcal{P}_t$.
The goal of the proposed method is to optimize
\begin{equation}
    \mathcal{L}(\bm{W}) = \mathrm{arg} \mathop{\min}_{\bm{w}_t}\frac{1}{T} \sum_{t=1}^{T} \mathbb{E}_{(\mathcal{X}, \mathcal{A})\sim \mathcal{P}_t} [\mathcal{L}_t(\bm{w}_t; \mathcal{S}_t, \mathcal{D}_t)],
    \label{eq1}
\end{equation}
where $\bm{W}=\lbrace \bm{w}_1, \bm{w}_2, ..., \bm{w}_T\rbrace$ is the parameter set of the client models, and $\mathcal{S}_t$ represents the shared information from all clients except $c_t$. However, as the true underlying distribution is often inaccessible, the goal is generally achieved through empirical risk minimization. 
Consequently, the training objective can be computed as follows:
\begin{equation}
    \mathrm{arg} \mathop{\min}_{\bm{w}_t} \sum_{t=1}^{T} [\mathcal{L}_{\mathrm{CE}} + \mathcal{R} + \mathcal{L}_{\mathrm{d}}],
\end{equation}
where $\mathcal{L}_{\mathrm{CE}}$ is a local average loss over personal training data, specifically, cross-entropy loss, $\mathcal{R}$ serves as a regularization term applied during the training process to mitigate overfitting among clients, and  $\mathcal{L}_{d}$ represents a distance loss to ensure the shared information can be utilized efficiently. The proposed method enables the simultaneous training of personalized VQA models tailored for diverse data distributions while sharing information instead of private data, thereby ensuring privacy and facilitating mutual benefits among clients.

\vspace{-1em}
\subsection{Prompt-based Visual Question Answering Model}
\label{client}
\begin{figure}[t]
\centering   
\includegraphics[width=8.5cm]{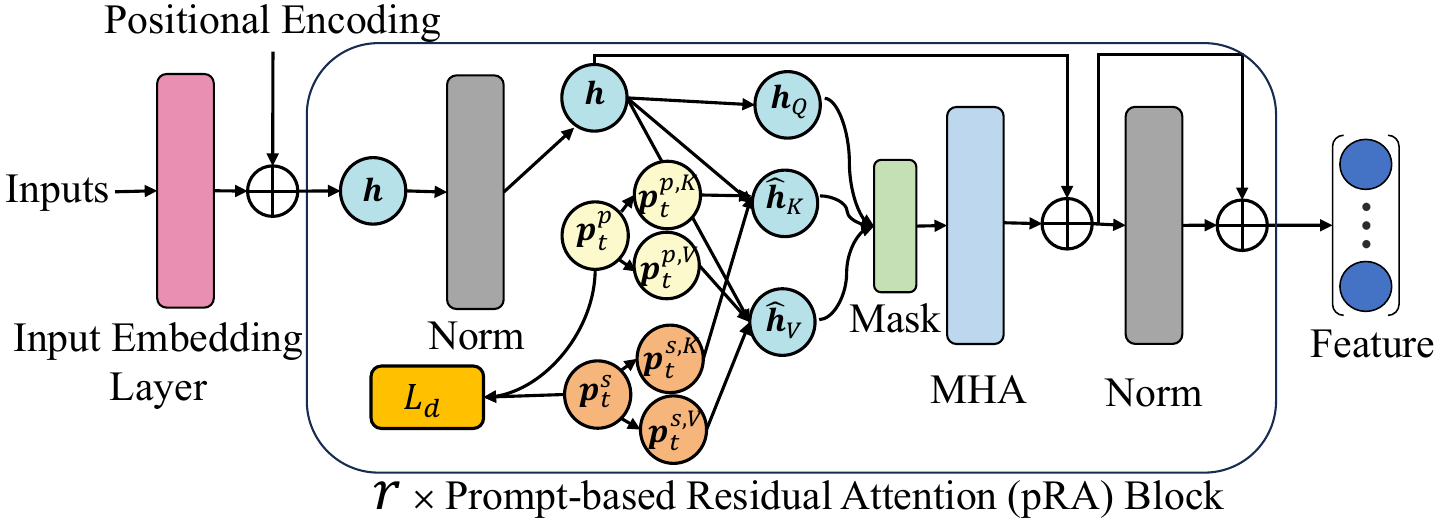}
\caption{The proposed transformer encoder that introduces the prompt-based self-attention block. Local prompt $\bm{p}_t^p$ and shared prompt $\bm{p}_t^s$ are integrated into the process of self-attention, and the distance between them is controlled by the distance loss $\mathcal{L}_{d}$.}
\label{fig:pRA}
\vspace{-1.5em}
\end{figure}
We describe the proposed prompt-based VQA client model with a general client $c_t$ as an example. We employ a specific variational auto-encoder structure~\cite{kingma2013auto} that integrates an image encoder, a text encoder, and an answering decoder, where the two encoders have the same transformer architecture and different parameters.
The original transformer encoder consists of an input embedding layer and a set of multi-head self-attention (MHA) based residual attention (RA) blocks. The RA block uses self-attention mechanisms and feedforward neural networks, combined with residual connections and normalization, to effectively process data sequences.

Directly using the original architecture can lead to excessive load in the communication process, significantly reducing the efficiency of federated learning.
Therefore, we propose a novel transformer encoder that introduces the prompt-based residual attention (pRA) block as shown in Fig.~\ref{fig:pRA}. The pRA block uses a set of learnable prompts as an additional input of the MHA layer.
Specifically, for client $c_t$, we introduce local prompt $\bm{p}_t^p \in \mathbb{R}^{r\times L\times D}$ trained locally and shared prompts $\bm{p}_t^s \in \mathbb{R}^{r\times L\times D}$ obtained from other clients, where $r$ is the number of the pRA blocks, and $L$ is the length of the prompts. In contrast to the original RA block, the proposed pRA block introduces the prompts into the MHA layer using Prefix-Tuning (Pre-T)~\cite{li2021prefix}. Pre-T splits $\bm{p}_t^p$ and $\bm{p}_t^s$ into $\lbrace \bm{p}_t^{p, K}, \bm{p}_t^{p,V}\rbrace \in \mathbb{R}^{1\times L/2 \times D}$ and $\lbrace \bm{p}_t^{s,K}, \bm{p}_t^{s,V}\rbrace \in \mathbb{R}^{1\times L/2 \times D}$ for each of the $r$ blocks, respectively. 

Each input image or question is embedded by the input embedding layer to $\bm{h} = \lbrace \bm{h}_Q, \bm{h}_K, \bm{h}_V\rbrace \in \mathbb{R}^{D}$, where the positional encoding has been added. Next, $h$ is fed into the MHA layer, which is calculated as follows:
\begin{equation}
    \mathrm{MHA}(\bm{h}_Q, \bm{h}_K, \bm{h}_V, \bm{p}_t^p, \bm{p}_t^s) = \mathrm{Concat}(\hat{h}_1,...,\hat{h}_i, ....,\hat{h}_m)\bm{W}^O,
\end{equation}
\begin{equation}
    \hat{h}_i = \mathrm{Attention}(\bm{h}_Q\bm{W}_i^Q,\hat{\bm{h}}_K\bm{W}_i^K, \hat{\bm{h}}_V\bm{W}_i^V),
\end{equation}
\begin{equation}
    \hat{\bm{h}}_K=\left[\bm{p}_t^{p,K} \oplus \bm{p}_t^{s,K};\bm{h}_K\right],
     \hat{\bm{h}}_V=\left[\bm{p}_t^{p,V} \oplus \bm{p}_t^{s,V};\bm{h}_V\right].
\end{equation}
where $\bm{W}^O$, $\bm{W}_i^Q$, $\bm{W}_i^K$ and $\bm{W}_i^K$ are the parameter matrices of the transformer, and $\mathrm{Attention}(\cdot)$ is the self-attention. 
Moreover, we merge the original mask of the MHA layer with a all-ones matrix as the new mask to ensure that each block can refer to all the prompts during the self-attention process. In particular, we do not add the location information for the prompts because they are introduced directly as parameters.

The outputs of the last pRA block of the image encoder and the text encoder are the image and language features of the input image $I_t^j$ and question $Q_t^j$. Next, the extracted features are joined and input into the personalized answer layer consisting of an MLP to output the probabilities of the predicted answer ${\hat{A}_t^j}$.
We employ the cross-entropy loss to train the model, which can be calculated as
\begin{equation}
    \mathcal{L}_{CE} = -\sum_{j=1}^{N_t} A_t^j \log(\hat{A}_t^j).
\end{equation}
Moreover, to prevent heterogeneity between local and shared prompts leading to degradation of the results, we also introduce the following loss to control the distance between the two prompts:
\begin{equation}
    \mathcal{L}_{d} = 1 - \frac{\bm{p}_t^p \cdot \bm{p}_t^s}{\|\bm{p}_t^p\|_2 \times \|\bm{p}_t^s\|_2},
\end{equation}
where \( \|\bm{p}_t^p\|_2 \) and \( \|\bm{p}_t^s\|_2 \) are the L2 norms of the \( \bm{p}_t^p \) and \( \bm{p}_t^s \), respectively.
The final loss can be calculated as follows:
\begin{equation}
    \mathcal{L}_\mathrm{client} = \mathcal{L}_{CE} + \alpha\mathcal{L}_{d} + \beta\mathcal{R},
\end{equation}
where $\mathcal{R} = \|\bm{w}_t^\mathrm{mlp}\|_2^2$ is the L2 regularization constraint on the weights of the MLP layer to prevent overfitting.

All clients use parameter matrices with the same weights while training their respective prompts during the training process. By introducing the pRA block, the prompts can be used for client-to-client communication, which reduces the computational complexity by several orders of magnitude.
\vspace{-1em}
\subsection{Client-to-Client Communication}
\begin{figure}[t]
\centering
\includegraphics[width=8cm]{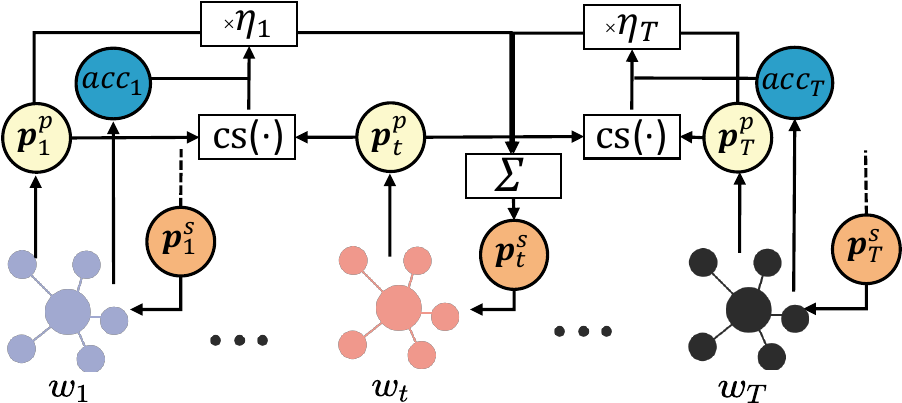}
\caption{The proposed client-to-client communication process. Each client communicates to update $\bm{p}_t^p$ and get $\bm{p}_t^s$ from the other clients. The reliability parameter $\eta$ controls the weights of different clients in the communication process, which is computed from the accuracy $acc$ and the similarity function $\mathrm{cs}(\cdot)$.}
\label{fig:c2c}
\vspace{-2em}
\end{figure}
We decouple the VQA model of client $c_t$ into the shared feature extractors and the personalized final answer layer, which are parameterized by $\bm{p}_t^p$ and $\bm{w}_t^\mathrm{mlp}$, respectively.
Each client has a task similar to multitask learning, where the number of the answer list is completely different between clients. The clients share the information through the shared prompt $\bm{p}_t^s$, and Eq.~(\ref{eq1}) can be further presented as follows:
\begin{equation}
 \mathcal{L}(W) = \mathrm{arg} \mathop{\min}_{\bm{p}_t^p, \bm{w}_t^\mathrm{mlp}}\frac{1}{T} \sum_{t=1}^{T} \mathbb{E}_{(\mathcal{X}, \mathcal{A})\sim \mathcal{P}_t} [\mathcal{L}_t(\bm{p}_t^p, \bm{w}_t^\mathrm{mlp};\bm{p}_t^s, \mathcal{D}_t)].
 \label{eq8}
\end{equation}

Since there are differences in the performance and tasks of clients, there should be a tendency to access shared information on the client. Therefore, we introduce a set of client's reliability parameter $\{\eta_t\}_{t=1}^{T}$ obtained from the performance and task similarity. For client $c_t$, the shared prompt $\bm{p}_t^s$ can be calculated as follows:
\begin{equation}
    \bm{p}_t^s = \sum_{{i=1, i \neq t}}^{T} \eta_i \bm{p}_i^p,
\end{equation}
\begin{equation}
    \eta_i = \frac{acc_i\times \mathrm{cs}(\bm{p}_i^p, \bm{p}_t^p)}{\sum_{{j=1, j \neq t}}^{T} acc_j\times \mathrm{cs}(\bm{p}_j^p, \bm{p}_t^p)},
    \label{eta}
\end{equation}
where $acc_i$ denotes the current accuracy of the client $c_i$ in answering the question, and $\mathrm{cs}(\cdot)$ denotes cosine similarity. 
It can be seen from Eq.~(\ref{eta}) that clients with high accuracy and higher similarity to the target client $c_t$ have higher reliability in the communication process.
Therefore, introducing the reliability parameter makes clients refer more to the information from clients with high accuracy, similar to their tasks, preventing the interference of invalid information. We provide an overview of the client-to-client communication in Fig.~\ref{fig:c2c}.

After several local iterations, the client $c_t$ interacts with other clients to update the shared information before making local updates. By sharing parameters between clients, the client $c_t$ can access additional information without changing the end-to-end training process that only accesses the private data. The main training procedure is summarized in Algorithm~\ref{alg1}.
\begin{algorithm}[t]
    \caption{Prompt-based pFL method for medical VQA} 
    \label{alg3} 
    \begin{algorithmic}[1]
 \REQUIRE $I$-Number of iterations, $I_L$-Number of local iterations, \\$\alpha$ — learning rate, $T$-Number of clients
 \STATE Initialization: client set $C=\{\bm{p}_t^p, \bm{p}_t^s, \mathbf{w}_t^{\mathrm{mlp}}\}_{t=1}^T$
 \FOR{$i=1$ to $I$}
     \FOR{$t=1$ to $T$}
  \FOR{$j=1$ to $I_L$}
      \STATE $\bm{p}_t^p = \bm{p}_t^p - \alpha\nabla_{\bm{p}_t^p}(\mathcal{L}_{\mathrm{client}}+\mathcal{L}_{d})$
      \STATE $\bm{w}_t^{\mathrm{mlp}} = \bm{w}_t^{\mathrm{mlp}} - \alpha\nabla_{\bm{w}_t^{\mathrm{mlp}}}\mathcal{L}_{\mathrm{client}}-\beta\nabla_{\bm{w}_t^{\mathrm{mlp}}}\|\bm{w}_t^\mathrm{mlp}\|_2^2$
  \ENDFOR
  \STATE $\bm{p}_t^s = \sum_{\substack{i=1,\ i \neq t}}^{T} \eta_i \bm{p}_i^p$
     \ENDFOR       
 \ENDFOR
 \ENSURE $C=\{\bm{p}_t^p, \bm{p}_t^s, \bm{w}_t^{\mathrm{mlp}}\}_{t=1}^T$
    \end{algorithmic}
    \label{alg1}
\end{algorithm}
\vspace{-1em}
\section{EXPERIMENTS}
\vspace{-1em}
\subsection{Settings}
%
$\mathbf{Dataset\ and\ Client}$.
In our experiments, we leveraged medical images and closed-type QA pairs from both VQA-RAD~\cite{lau2018dataset} and Slake~\cite{liu2021slake} datasets. These datasets were further partitioned into sub-datasets based on the anatomical focus of the images. We split the datasets into sub-datasets depending on the organ of the image taken and set up client models for each sub-dataset. 
Finally, we set four clients for Slake including ``Lung'', ``Abdomen'', ``Brain'' and ``Others'', which contain 734, 829, 461, and 335 question pairs, respectively. 
We set three clients for VQA-RAD including ``Chest'', ``Abdomen'' and ``Brain'', which contain 511, 439, and 269 question pairs, respectively. 
We adhered to the original dataset division ratios as 70\% for training, 15\% for validation and 15\% for testing in Slake, and 85\% for training and 15\% for testing in VQA-RAD.

Our architecture employs the Contrastive Language–Image Pre-training (CLIP)~\cite{radford2021learning} model as the backbone for both the image and text encoders. The answer layer is implemented as a 2-layer MLP with a hidden size of 512 and a dropout rate of 0.2. We standardized the length of all prompts to four. The weight $\alpha$ of the loss was set to 0.5, and $\beta$ was set to 0.001. During the training, we set the global epoch to 30 and each local iteration to 5. Based on the experimental results, we used a uniform learning rate of 0.01 for all client models, which is multiplied by 0.1 every ten epochs. After adjusting for the block number $r$, we found that the model had the best performance when $r$ was set to 12.

$\mathbf{Comparison\ Methods}$. 
We compared the proposed method with several state-of-the-art medical VQA models.
We used the DNNs-based methods, VGG+SAN~\cite{liu2021slake} and MEVF~\cite{10.1007/978-3-030-32251-9_57} for Slake and VQA-RAD, respectively. And we used the transformer-based methods, PMCLIP~\cite{eslami2021does} and M2I2~\cite{li2023self} for both of them. To ensure consistent experimental conditions, we trained the comparison separately using split sub-datasets and did not use pre-training.

$\mathbf{Ablation\ Studies}$. 
We used ablation experiments to demonstrate the effectiveness of the proposed method.
We separately experimented with the proposed method with the following settings. AS1: w/o prompt and pFL, AS2: w/o pFL,
AS3: w/o image prompt, AS4: w/o text prompt.
\vspace{-1em}
\subsection{Results}
\begin{table*}[t]
\centering
\begin{minipage}[t]{0.47\linewidth}
    \caption{The quantitative results of each client of Slake.}
    \centering
    \begin{tabular}{ccccc}
    \hline
    & Brain& Abdomen& Others & Lung \\ \hline
    VGG+SAN~\cite{liu2021slake} & 76.92         & 72.34              & 74.41  & 75.89 \\ \hline
    PMCLIP~\cite{eslami2021does}& 79.12         & \textbf{80.85}     & 76.74  & 82.98 \\ \hline
    M2I2~\cite{li2023self}      & 79.12         & 68.09              & 74.42  & 81.56 \\ \hline\hline
    AS1& 79.12                     & 70.92             & \underline{81.40}  & \underline{83.69}           \\ \cline{1-5}
    AS2& \textbf{81.32}            & 72.34             & 79.07              & 82.98                \\ \cline{1-5}
    AS3& \underline{80.22}         & \underline{73.76} & \textbf{81.84}     & 82.27                \\ \cline{1-5}
    AS4& \underline{80.22}         & 70.92             & 76.74              & \textbf{84.40}        \\ \cline{1-5}
    PM & 76.92                     & 72.34             & \underline{81.40}  & \textbf{84.40}       \\ \hline
    \end{tabular}
    \label{res-slake}
\end{minipage}
\hfill
\begin{minipage}[t]{0.5\linewidth}
    \caption{The quantitative results of each client of VQA-RAD.}
    \centering
    \begin{tabular}{cccc}
    \hline
    & Brain & Abdomen & Chest               \\ \hline
    MEVF~\cite{10.1007/978-3-030-32251-9_57} & 68.63 & 66.67 & 60.61                                    \\ \hline
    PMCLIP~\cite{eslami2021does} & \underline{73.53} & \underline{70.00} & 62.12                                     \\ \hline
    M2I2~\cite{li2023self} & 64.38 & \textbf{71.11} & \underline{63.64} \\ \hline\hline
    AS1& \textbf{74.24} & 67.78          & 61.76                                \\ \hline
    AS2& 68.18          & 63.33          & 61.76                                                    \\ \hline
    AS3& \textbf{74.24} & 67.78          & 61.76                                                    \\ \hline
    AS4& 69.70          & 67.78          & 60.78                                                    \\ \hline
    PM & \textbf{74.24} & 68.89 & \textbf{64.71}                                           \\ \hline
    \end{tabular}
    \label{res-VQARAD}
\end{minipage}
\vspace{-1em}
\end{table*}
\begin{figure}[t]
  \centering
  \subfigure[Slake]{
    \includegraphics[width=0.45\linewidth]{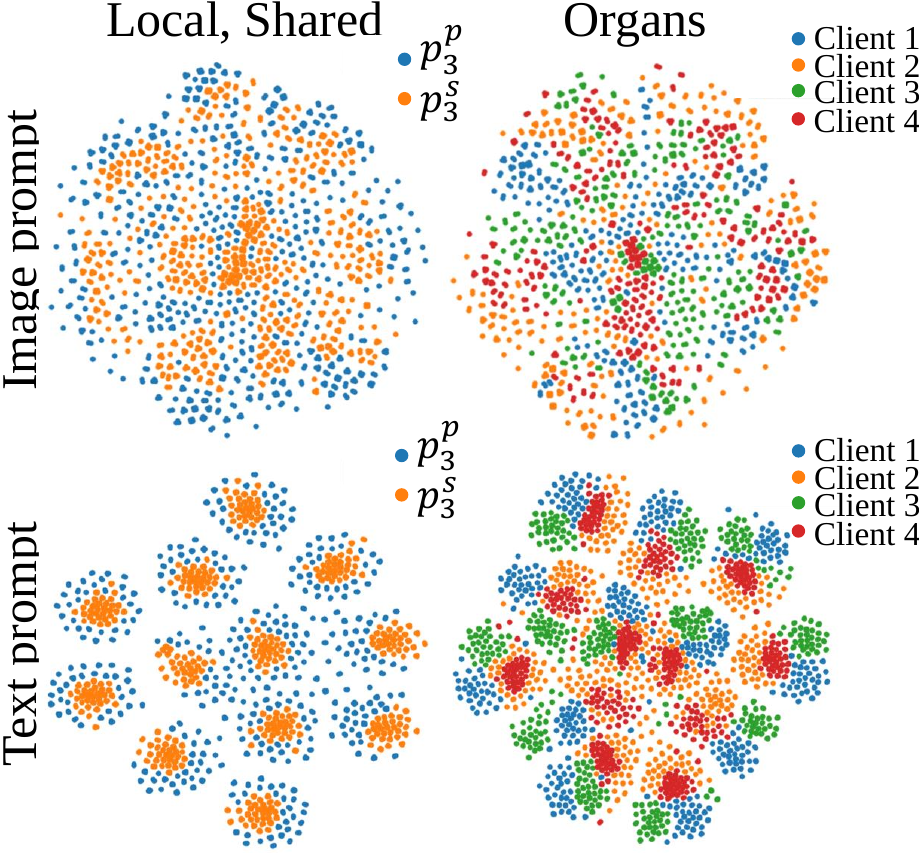}
    \label{fig:subfig1}
  }
  \subfigure[VQA-RAD]{
    \includegraphics[width=0.45\linewidth]{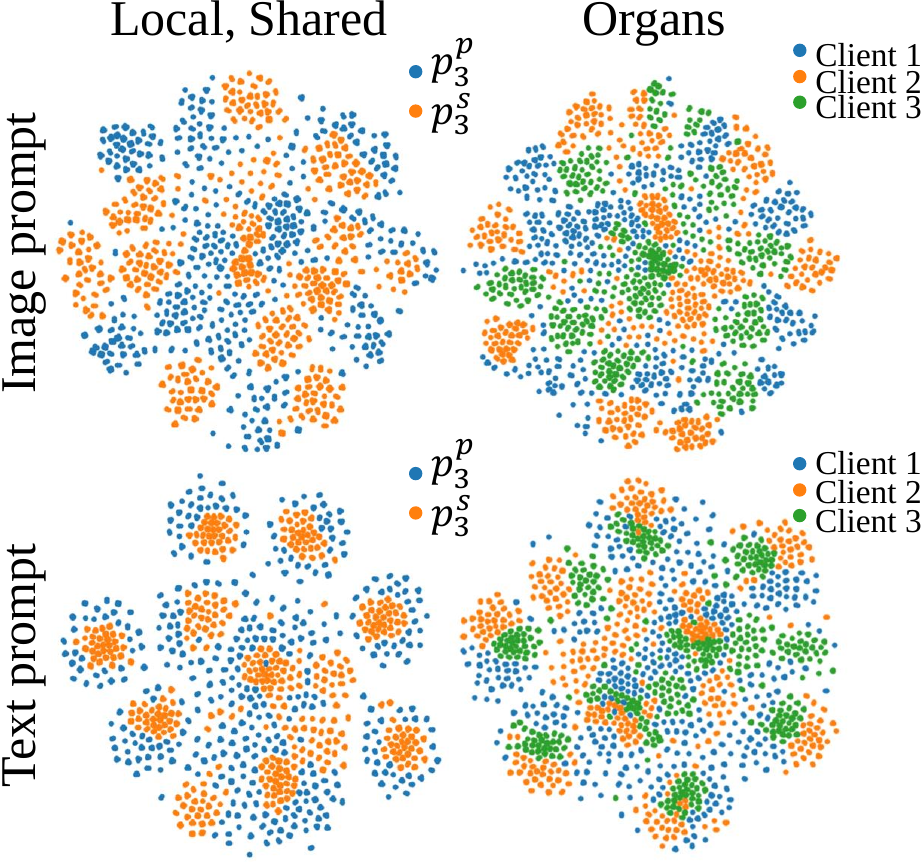}
    \label{fig:subfig2}
  }
  \caption{The tSNE visualization of the prompt embedding results on the Slake and VQA-RAD datasets. The ``Local, Shared'' represents the local and shared prompts in client 3 that is trained in the sub-dataset of the Abdomen. The ``Ograns'' represents the local prompt of each client.}
  \label{fig:main}
  \vspace{-1.5em}
\end{figure}
\begin{figure}[t]
\centering  
\includegraphics[width=8cm]{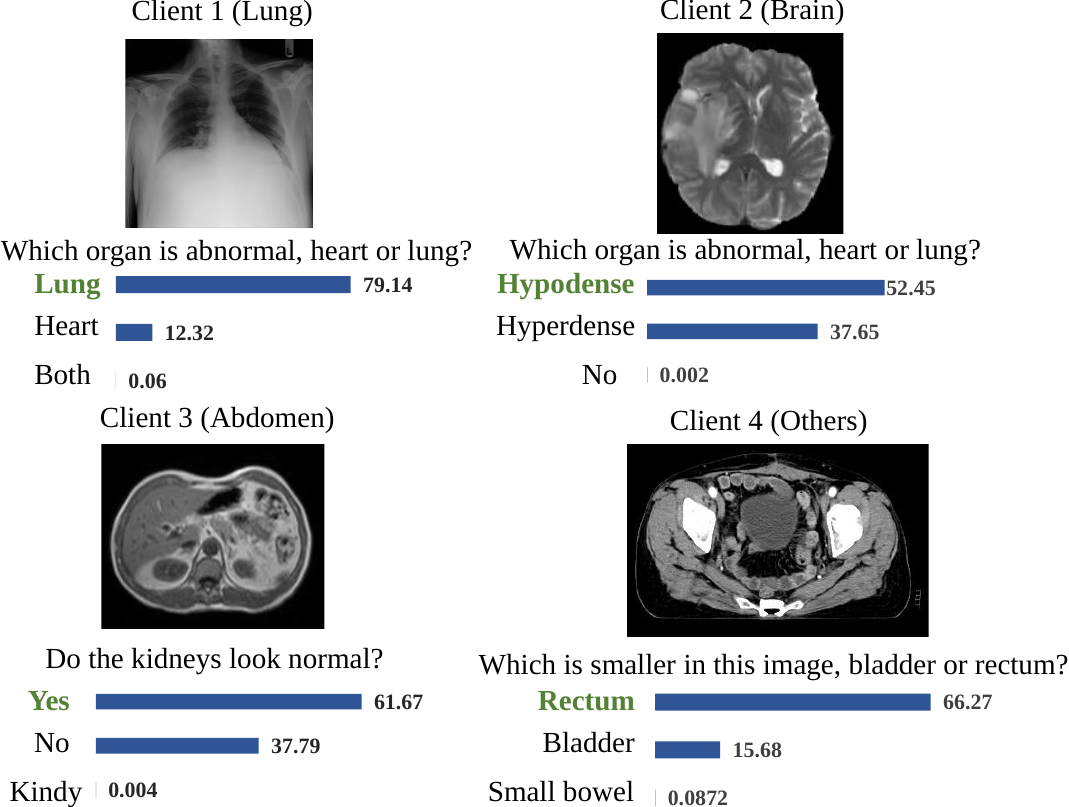}
\caption{Samples of the generated answers on Slake. The results show the four clients we set up, with the corresponding sub-datasets in parentheses.}
\label{fig:res}
\vspace{-1.5em}
\end{figure}
We conducted experiments on the accuracy of different client models in answering questions, which are shown in Tables~\ref{res-slake} and~\ref{res-VQARAD}.
The high accuracy of PM, as well as AS3 and AS4, which incorporated pFL, attest to the efficacy of the pFL method.
Compared to AS1, AS2 does not exhibit significant improvements, demonstrating that incorporating prompts offers limited enhancements to the model's performance.
Moreover, for Slake, the text prompt had a more positive impact than the image prompt, and for the VQA-RAD, both the image and text prompts improved the accuracy of the model. 

We also visualized the prompts with tSNE~\cite{van2008visualizing}. From the visualization results in Fig.~\ref{fig:subfig1}, consistent with the quantitative results, we see that the text prompt of the Slake is more accurately mapped to different regions, while the image prompt is relatively cluttered. Furthermore, as shown in both Figs.~\ref{fig:subfig1} and~\ref{fig:subfig2}, the 12 regions in the figures correspond to the 12 pRA blocks in the proposed method, which can prove that each block contains different potential information. The clear separation into several parts within each region in the ``Organ'' can prove that the prompts trained under different data distributions contain entirely different information. The ``Local, Shared'' results demonstrate the effectiveness of the proposed distance loss, where the corresponding local and shared prompts in each block are mapped to a similar space.
We also qualitatively evaluated the model answering the questions as shown in Fig.~\ref{fig:res}. It can be seen that our model can correctly answer the questions corresponding to the medical images.
In the experiments, the parameters to be updated of the proposed method were only 0.05\% of that of previous pFL methods.
\vspace{-1em}
\section{CONCLUSION}
\label{sec:typestyle}
The proposed prompt-based personalized federated learning method for medical VQA can train personalized models for sub-datasets of medical data with different distributions, which overlooked problems of medical data heterogeneity and privacy protection. By using the prompt-based self-attention block, the client model circumvented the computational intricacies of traditional model weight sharing. Both qualitative and quantitative results show that our method allows the model to gain in communication. Moreover, the proposed method is expected to be used in any transformer-based VQA baseline models.


\newpage
\bibliographystyle{IEEEbib}
\bibliography{refs}
\end{document}